\documentclass{article}

\usepackage{CJKutf8}
\usepackage{longcat_style}
\usepackage{adjustbox}
\usepackage[utf8]{inputenc} %
\usepackage[T1]{fontenc}    %
\usepackage{newunicodechar}
\usepackage{hyperref}       %
\usepackage{xcolor}
\usepackage[normalem]{ulem} %
\hypersetup{
    colorlinks=true,      %
    linkcolor=blue,      %
    urlcolor=blue,       %
    citecolor=blue,      %
    linkbordercolor=blue, %
    urlbordercolor=blue,
    citebordercolor=blue,
    pdfborderstyle={/S/U/W 1}, %
}
\usepackage{float}
\usepackage{url}            %
\usepackage{booktabs}       %
\usepackage{amsfonts}       %
\usepackage{nicefrac}       %
\usepackage{microtype}      %
\usepackage{lipsum}		%
\usepackage{graphicx}
\usepackage{natbib}
\usepackage{doi}
\usepackage{amsmath}
\usepackage{amssymb} %
\usepackage{xspace}
\usepackage{enumitem}
\usepackage{multirow}
\usepackage{subcaption} 
\usepackage{makecell}
\usepackage{hyperref, cleveref}
\usepackage{pifont}
\usepackage[inkscapelatex=false]{svg}
\usepackage{caption}
\usepackage[most]{tcolorbox}
\usepackage[table,xcdraw]{xcolor}  
\DeclareCaptionLabelSeparator{pipe}{ | }
\usepackage{tcolorbox}
\usepackage{wrapfig}
\newtcolorbox{coloredquote}[1][]{
    colback=green!5!white,  
    colframe=green!70!black, 
    boxrule=2pt,
    arc=7pt,
    left=6pt,
    right=6pt,
    top=4pt,
    bottom=4pt,
    title=#1
}

\tcbset{
    mybox/.style={
        colback=gray!10, 
        colframe=black, 
        coltitle=white, 
        fonttitle=\bfseries, 
        boxrule=0.75mm, 
        arc=1mm, 
        leftrule=2mm, 
        width=\linewidth, 
        boxsep=5pt, 
        left=10pt, 
        right=10pt, 
    }
}

\definecolor{gemblue}{RGB}{225, 240, 255}

\setlist[itemize]{leftmargin=*}
\setlist[enumerate]{leftmargin=*}
\setlist[description]{leftmargin=*}

\title{Unlocking Implicit Experience: \\ Synthesizing Tool-Use Trajectories from Text}

\author{
    Zhihao Xu\textsuperscript{1}\textsuperscript{2}\thanks{Equal Contributions.}, 
    Rumei Li\textsuperscript{2}\footnotemark[1], 
    Jiahuan Li\textsuperscript{2}, \\
    \textbf{Rongxiang Weng}\textsuperscript{2}\footnotemark[2], 
    \textbf{Jingang Wang}\textsuperscript{2}\thanks{Corresponding authors.}, 
    \textbf{Xunliang Cai}\textsuperscript{2}, 
    \textbf{Xiting Wang}\textsuperscript{1} \\
    \textsuperscript{1}Renmin University of China, 
    \textsuperscript{2}Meituan, China \\
    \texttt{zhihaoxu@ruc.edu.cn}
}

\renewcommand{\headeright}{\raisebox{-0.2\height}{\includegraphics[height=1.8em]{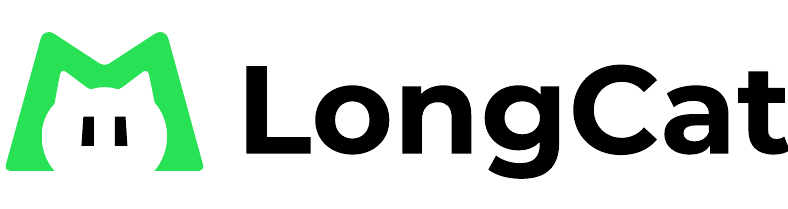}}}
\renewcommand{\shorttitle}{Unlocking Implicit Experience: Synthesizing Tool-Use Trajectories from Text}

\begin{document}
\maketitle
\setcounter{footnote}{0}

\begin{abstract}
Enabling Large Language Models (LLMs) to effectively utilize tools in multi-turn interactions is essential for building capable autonomous agents. However, acquiring diverse and realistic multi-turn tool-use data remains a significant challenge. 
In this work, we propose a novel text-based paradigm. We observe that textual corpora naturally contain rich, multi-step problem-solving experiences, which can serve as an untapped, scalable, and authentic data source for multi-turn tool-use tasks. Based on this insight, we introduce GEM, a data synthesis pipeline that enables the generation and extraction of multi-turn tool-use trajectories from text corpora through a four-stage process: relevance filtering, workflow \& tool extraction, trajectory grounding, and complexity refinement. To reduce the computational cost, we further train a specialized Trajectory Synthesizer via supervised fine-tuning. This model distills the complex generation pipeline into an efficient, end-to-end trajectory generator. 
Experiments demonstrate that our GEM-32B achieve a 16.5\% improvement on the BFCL V3 Multi-turn benchmark. Our models partially surpass the performance of models trained on $\tau$-bench (Airline and Retail) in-domain data, highlighting the superior generalization capability derived from our text-based synthesis paradigm.
Notably, our Trajectory Synthesizer matches the quality of the full pipeline while significantly reducing inference latency and costs.
\end{abstract}
\section{Introduction}
The pursuit of Artificial General Intelligence (AGI) relies on the development of autonomous agents capable of perceiving, reasoning, and acting in complex real-world environments~\citep{team2025kimi, liu2025deepseek, team2025longcat}. Such agents are required to dynamically utilize diverse tools to extend their capabilities and accomplish multi-step tasks. While large language models (LLMs) have demonstrated impressive tool-use capabilities, they still struggle in realistic multi-turn interactions, particularly when faced with ambiguous instructions, long-context dependencies, and unexpected errors~\citep{patilberkeley}. These limitations constrain their practical application in agent-based systems~\citep{yao2024tau, barres2025tau}. 

\begin{figure}[]
    \centering
    \includegraphics[width=\textwidth]{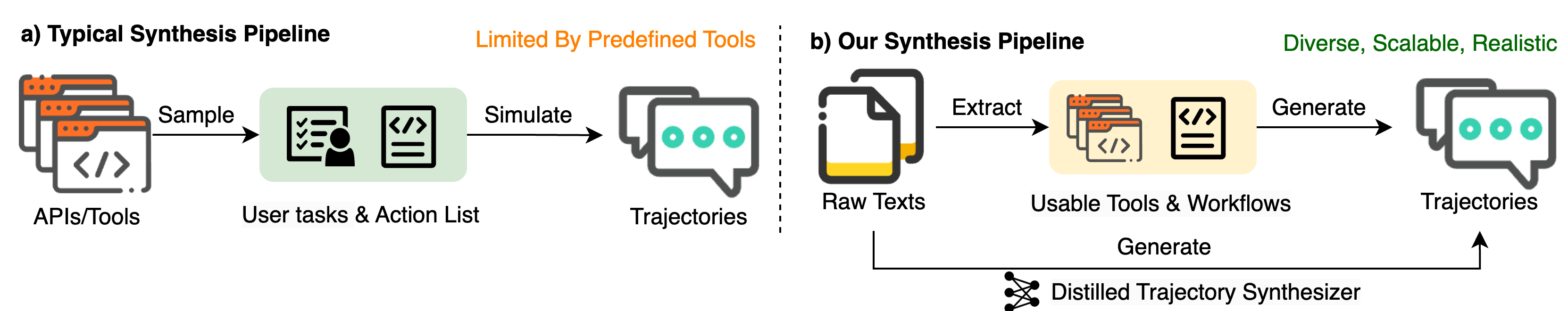}
    \caption{We propose the "Text to Trajectory" paradigm, instead of relying on predefined tools to generate tasks and trajectories.}
    \label{fig:paradigm}
\end{figure}

The primary bottleneck in training autonomous agents lies in the scarcity of high-quality, multi-turn tool-use trajectories, which are rarely found in real-world scenarios. 
To overcome this data scarcity, current research favors a tool-centered simulation paradigm (Fig. \ref{fig:paradigm}). These methods typically rely on predefined API sets to synthesize user tasks and simulate interactions~\citep{qin2023toolllm, liu2024apigen, guo2024stabletoolbench, zeng2025toolace, prabhakar2025apigen, liu2024toolace, yin2025magnet}. However, gathering a sufficiently diverse and comprehensive tool set is inherently expensive and difficult. The resulting tool-use training data is often limited by the scope of the predefined APIs, while the ultimate goal of agentic training is the exposure to a sufficiently broad range of scenarios during training to enable agents to generalize effectively to unseen environments and scenarios~\citep{fang2025towards}. This raises a critical question: \textit{Can we bypass the dependency on predefined tools and synthesize more diverse, high-quality trajectories directly from real-world?}

In this paper, we answer this question by proposing a novel \textbf{text-based extraction paradigm}. We observe that the raw text corpora used for pre-training large language models inherently contain multi-turn tool-use patterns. Although such texts do not contain explicit agentic trajectories, they often document rich, real-world mutli-step problem-solving experiences, which can be extracted and transformed into multi-turn tool-use data (e.g., "the procedure for hospital reimbursement claims"). Our preliminary analysis of Ultra-fineweb~\citep{wang2025ultra} confirms that such texts contain actionable logical sequences spanning diverse domains, indicating that unstructured text serves as an untapped, scalable, and authentic source for synthesizing agentic training data.

To operationalize this paradigm, we introduce \textsc{GEM}, a data synthesis pipeline that enables generation and extraction of multi-turn tool-use trajectory from text corpora. The GEM pipeline proceeds in four stages: (1) \textbf{Selection}, identifying text segments rich in multi-step workflows; (2) \textbf{Extraction}, deriving structured workflows and tool definitions to map human-described procedures to agent actions; (3) \textbf{Generation}, employing a powerful model to convert text and abstract workflows into concrete user-agent interactions; and (4) \textbf{Refinement}, enhancing the complexity and diversity of the trajectories followed by rigorous verification.

Extensive experiments validate the effectiveness of our approach. When fine-tuned on data synthesized by our pipeline, the GEM-32B model achieves a 14.9\% improvement on BFCL V3 Multi-Turn benchmark. 
Notably, We find that out-of-domain training data generated through our paradigm achieve performance on $\tau^2$-bench that is comparable to models trained on $\tau$-bench in-domain data, thereby demonstrating the strong generalization capability of our approach.
Building upon this high-quality data, we further develop a specialized Trajectory Synthesizer to internalize the "text-to-trajectory" mapping. This synthesizer provides a cost-effective, end-to-end solution for large-scale data generation, matching the quality of the multi-stage synthesis pipeline while significantly reducing costs.

In summary, our contributions are threefold:
\begin{itemize}
\item We propose a new paradigm for agent trajectory synthesis that directly extracts multi-turn tool-use trajectories from text corpora. This paradigm unlocks an untapped, scalable source of authentic human problem-solving behaviors for training autonomous agents.

\item We develop GEM, a data synthesis pipeline that transforms raw text into multi-turn tool-use trajectories to prove the effectiveness of our proposed paradigm. We use GEM to synthesize a high-quality dataset that leads to significant performance gains: 13.8\% on $\tau^2$-bench and 16.5\% on BFCL V3 Mutli-Turn.

\item To enable cost-effective and scalable data generation, we develop a specialized Trajectory Synthesizer via supervised fine-tuning. We demonstrate that this model successfully distills the full GEM pipeline into an end-to-end generator, matching the original synthesis quality while significantly reducing costs.
\end{itemize}

\section{Problem Formulation}

In this paper, we address the challenge of acquiring general multi-turn tool-use agent trajectories. We propose a novel paradigm that synthesizes such trajectories by extracting tool-use workflows directly from large-scale textual corpora. 
This approach naturally enjoys these advantages: the scale of real-world text ensures scalability, its inherent diversity covers multiple domains, and its foundation in grounded human problem-solving experiences yields high-quality, realistic agent data. 
Formally, we define the paradigm as:
\begin{itemize}
    \item \textbf{Input:} Let $\mathcal{C}$ denote a large-scale text corpus, where each element $c_i \in \mathcal{C}$ is a raw, unstructured text segment (e.g., a document or narrative) that contains multi-step workflows.

    \item \textbf{Output:} A complete list of tools $\mathcal{P} = \{p_1, \dots, p_m\}$, where $m$ denotes the number of available tools and each $p_i$ represents a tool defined in the standard OpenAI format. Additionally, a structured multi-turn tool-use trajectory $T = \{s,(u_1, a_1, o_1), \dots, (u_n, a_n, o_n)\}$ is produced, where $s$ denotes the system prompt and at each turn $t$, $u_t$ corresponds to a user's query or request, $a_t$ denotes the assistant's natural language response or tool call, and $o_t$ is the resulting observation from the tool's execution.
\end{itemize}

\section{Methodology}
In this section, we introduce \textsc{GEM}, an agentic synthesis pipeline designed to automatically extract and generate multi-turn tool-use trajectories directly from large-scale text corpora. 
We first conduct a preliminary analysis of text corpora in Section~\ref{subsec:pre_analysis}.
Subsequently, we detail our trajectory collection pipeline in Section~\ref{subsec:data_coll}.

\subsection{Preliminary Analysis}
\label{subsec:pre_analysis}


\begin{figure}[htbp]
    \centering
    \begin{minipage}{0.48\textwidth}
        \centering
        \includegraphics[width=0.5\linewidth]{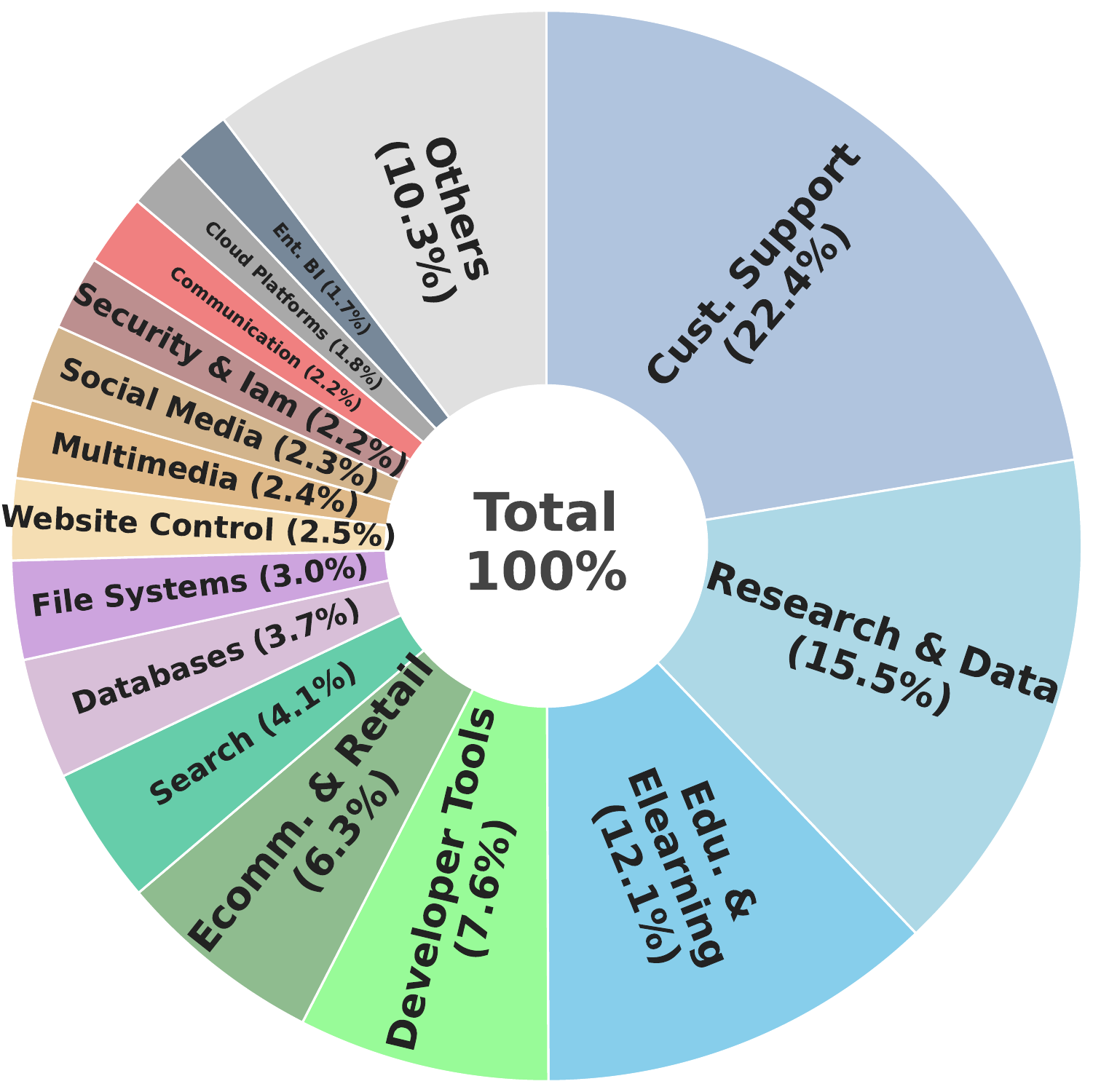}
        \caption{Distribution of the extracted task category from raw documents, showing the diversity of tasks covered in the text corpora.}
        \label{fig:domain_analysis}
    \end{minipage}
    \hfill
    \begin{minipage}{0.48\textwidth}
        \centering
        \includegraphics[width=0.9\linewidth]{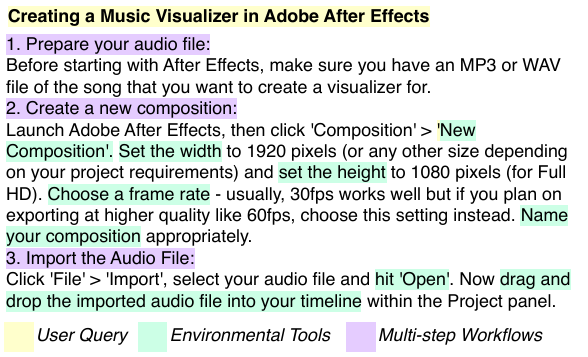}
        \caption{Preliminary case study.}
        \label{fig:pre_analysis_case}
    \end{minipage}
\end{figure}

\paragraph{Setup}
To assess the feasibility of our proposed paradigm, we first perform a preliminary analysis on the characteristics of large-scale unstructured data, using the Ultra‑fineweb corpus~\citep{wang2025ultra} as our main data source. We randomly sample approximately 250,000 raw text segments. Each segment is then processed through a sequential labeling pipeline: we first use a classifier to determine whether the text contains multi‑step operational procedures, which helps us identify segments suitable for conversion into tool‑use trajectories. Only those segments identified as containing procedural content are subsequently annotated with rich metadata, including platform, domain, and task category, via \texttt{Qwen3-8B} to evaluate the diversity of the corpus.
We also conduct a preliminary case study.
The prompt templates used for metadata annotation are provided in Appendix~\ref{app:metadata}. 

\paragraph{Results}
First, by observing a large number of text cases, we find that unstructured textual data inherently contain the three core components essential for constructing agentic trajectories as shown in Figure~\ref{fig:pre_analysis_case}: (1) \textbf{User Queries}, which emerge naturally as goals or problems stated in the text; (2) \textbf{Environmental Tools}, whose descriptions, APIs, or functionalities are often embedded within explanatory or instructional contexts; and (3) \textbf{Multi-step Workflows}, manifesting as step-by-step procedures or operational narratives. 

We find that around 14\% of the sampled segments contain explicit multi‑step workflows, indicating a substantial reservoir of procedural knowledge. Given the massive scale of available text corpora, this represents a significant potential source for generating diverse agent trajectories.
Furthermore, the identified procedural segments cover a wide spectrum of task categories and application scenarios—including education \& E-learning, data analysis as illustrated in Figure~\ref{fig:domain_analysis}. This diversity confirms that textual sources can provide the necessary variety in tasks, tools, and environments required for robust agent training. The detailed domain distribution is further reported in Appendix~\ref{app:domain}.

These findings collectively demonstrate that generating multi‑turn tool‑use trajectories directly from unstructured text is not only feasible but also taps into a vast and largely underexplored repository of agentic trajectory.

\subsection{GEM Synthesis Pipeline}
\begin{figure*}[]
    \centering
    \includegraphics[width=\textwidth]{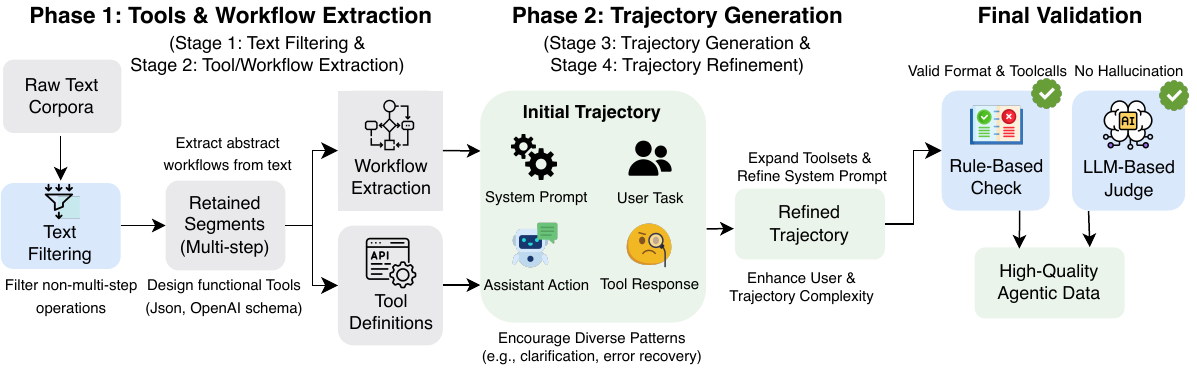}
    \caption{Overall Pipeline of GEM. The first phase extracts multi-step workflows from filtered text corpora and defines corresponding tools. The second phase generates and refines trajectories, incorporating diverse behavioral patterns and rigorous validation to produce high-quality training data.}
    \label{fig:pipeline}
\end{figure*}
\label{subsec:data_coll}
We now detail the data synthesis pipeline of \textsc{GEM} to generate and extract multi-turn tool-use data. The overall pipeline is illustrated in Figure~\ref{fig:pipeline}. All prompts used in this section are in Appendix~\ref{app:prompt}.

\paragraph{Stage 1: Text Filtering} 
To ensure the quality and realism of generated trajectories, the initial stage filters out raw text segments $c$ that do not describe multi-step operations. The filtering procedure uses the same annotation prompt and model as described in Section~\ref{subsec:pre_analysis}.

\paragraph{Stage 2: Workflow \& Tool Extraction} 
For each retained text segment $c$ containing multi-step operations, we proceed to extract structured abstract workflows and synthesize corresponding functional tools directly based on the text segment description. 
First, the model is instructed to identify all workflows and enumerate the individual steps within each (e.g., \texttt{search\_items} and then \texttt{edit\_item}). Importantly, the model is encouraged to recognize workflow complexity, including sequential dependencies, conditional logic, and uniqueness constraints, thereby enhancing the richness and practical relevance of the output. Concurrently, the model designs a set of API tools $\mathcal{P}$ in accordance with OpenAI schema standards to support the subsequent trajectory generation. Each tool $p$ is designed to perform a single, coherent function, with self-explanatory parameter names and well-specified data types. The final structured output comprises the abstract workflow descriptions alongside the complete tool definitions.

\paragraph{Stage 3: Trajectory Generation}
Based on the text, abstract workflows and tools synthesized in the previous stage, we now proceed to generate concrete multi-turn tool-use trajectories.
For each workflow and its corresponding set of tools, we employ a strong teacher model (\texttt{GLM-4.6} in this work) to generate a preliminary complete trajectory $T$. Our approach synthesizes the full trajectory directly in a single pass instead of simulating turn-by-turn conversations via a multi-agent system to ensure efficiency. Each generated trajectory include the following core components:

\begin{itemize}
    \item \textbf{System Prompt} $s$: Clear domain-specific rules extracted from the source text, establishing guidelines the assistant must follow throughout the conversation.

    \item \textbf{User Task} $(u_1,...u_n)$: A series of natural user requests that may be ambiguous or complex, reflecting real-world scenarios.
    
    \item \textbf{Assistant Responses} $(a_1,...,a_n)$: Demonstrations of intelligent problem-solving, strict adherence to domain rules, and correct tool invocation with appropriate parameters.
    
    \item \textbf{Tool Response} $(o_1,...o_n)$: Simulated tool outputs that are both complete and realistic, providing authentic feedback.
\end{itemize}
Here, $n$ denotes the number of conversational turns for each role.
To capture the diverse practical challenges inherent in real-world multi-turn interactions, we encourage the inclusion of various interaction patterns during generation. These include but are not limited to: refusing requests that exceed the assistant's capabilities, clarifying ambiguous user queries, and actively recovering from errors.

\paragraph{Stage 4: Refinement}
We observe that although the initial multi-turn dialogue trajectories $T$ are complete, they often lack sufficient complexity and tend to be relatively straightforward. To fully utilize and enhance the quality of these trajectories, we adopt a refinement strategy to further process them. Specifically, given $T$, we generate a refined trajectory $T'$ by expanding the variety of tools used, improving the realism of environmental responses, increasing the ambiguity and complexity of user requests, and ensuring the inclusion of non-trivial tool-call chains. We find this refinement essential for obtaining higher-quality agentic data, as demonstrated in Section~\ref{sec:ablation}.

\paragraph{Validation}
The trajectory filtering process integrates rule-based verification with LLM-based assessments to ensure high-quality outputs. Given a refined trajectory $T'$, we first apply a rule-based check to guarantee structural correctness. This involves verifying that all tools are correctly defined according to the OpenAI format and that each tool call corresponds to a valid function within the designated toolset, with argument names and types matching their definitions. We further validate the conversation format to confirm that tool responses meet all requirements and that all role tags are properly closed.
Beyond structural correctness, we employ an LLM-based judge (\texttt{Qwen3-32B} in this paper) to detect and eliminate hallucinations. This step specifically examines all tool calls to ensure that every generated parameter value is explicitly grounded in the dialogue context rather than being fabricated. Only those trajectories that successfully pass both validation stages are retained as the final set $T_{\text{final}}$ for use in supervised fine-tuning (SFT).

\paragraph{Data Synthesizer}
Generating such trajectories is costly and time-consuming. To address this, we propose training a data synthesizer via supervised fine-tuning (SFT), which learns a end-to-end mapping from text to multi-turn tool-use trajectories. This approach allows for the cost-effective synthetic training data.
For each data instance, the synthesizer takes as input $x$ an instruction ("Turn the following text into multi-turn tool-use trajectories") paired with a text segment. It then generates the corresponding output $y$, which includes both the necessary tool definitions and the resulting multi-turn tool-use trajectory. 
\label{subsec:data_synth}

\begin{table*}[t]
\centering
\resizebox{0.9\textwidth}{!}{%
\begin{tabular}{lccccc}
\toprule
\multirow{2}{*}{\textbf{Model}} & \textbf{Multi-Turn} & \textbf{Multi-Turn} & \textbf{Multi-Turn} & \textbf{Multi-Turn} & \textbf{Multi-Turn} \\
 & \textbf{Overall Acc} & \textbf{Base} & \textbf{Miss Func} & \textbf{Miss Param} & \textbf{Long Context} \\
\midrule
\multicolumn{6}{c}{\textit{Proprietary \& Large Scale Models}} \\
\midrule
GPT-4.1 & 38.88 & 47.50 & 32.50 & 32.50 & 43.00 \\
DeepSeek-V3.2-Exp & 37.38 & 41.50 & 39.50 & 33.50 & 35.00 \\
Gemini-2.5-Flash & 36.25 & 41.50 & 36.00 & 32.00 & 35.50 \\
\midrule
\multicolumn{6}{c}{\textit{8B Models}} \\
\midrule
Qwen3-8B & 18.00 & 24.00 & 17.00 & 13.50 & 17.50 \\
APIGEN-MT & 21.00 & 25.50 & 19.00 & 25.00 & 14.50 \\
TOUCAN & 21.88 & 27.00 & 21.50 & 21.00 & 18.00 \\
MUA & 21.13 & 29.50 & 18.00 & 20.00 & 17.00 \\
\rowcolor{gemblue} Qwen3-8B-GEM & \textbf{30.25} & \textbf{40.00} & \textbf{30.00} & \textbf{28.00} & \textbf{23.00} \\
\midrule
\multicolumn{6}{c}{\textit{32B Models}} \\
\midrule
Qwen3-32B & 28.35 & 34.00 & 24.50 & 25.50 & 29.50 \\
APIGEN-MT & 29.50 & 36.00 & 27.00 & 28.50 & 26.50 \\
TOUCAN & 35.00 & 41.00 & 37.50 & 26.00 & 35.50 \\
MUA & 26.25 & 30.00 & 25.00 & 23.00 & 27.00 \\
\rowcolor{gemblue} Qwen3-32B-GEM & \textbf{44.88} & \textbf{52.00} & \textbf{40.00} & \textbf{38.50} & \textbf{49.00} \\
\bottomrule
\end{tabular}%
}
\caption{\textbf{Performance comparison on the BFCL V3 benchmark.} The results are categorized by model scale. The best performance in each category is highlighted in \textbf{bold}. All metrics are reported as accuracy scores. The GEM-based models are highlighted in \colorbox{gemblue}{blue}.}
\label{tab:bfcl_results}
\end{table*}

\section{Experiments}

\subsection{Experimental Setups}
\noindent \textbf{Benchmark.} We evaluate our approach on two challenging benchmarks designed to assess multi-turn tool-use capabilities:
(1) BFCL V3~\citep{patilberkeley}: This benchmark involves agents interacting with a Python-based API environment. We focus specifically on the multi-turn scenarios, which are divided into four categories: \textit{Multi Turn Base}, \textit{Miss Param}, \textit{Miss Func}, and \textit{Long Context} with 200 tasks per category.
(2) $\tau^2$-bench~\citep{barres2025tau} (Airline and Retail): This benchmark evaluates user-agent interactions comprehensively within specialized real-world domains. We employ \texttt{GPT-4.1} as the user simulator Following the original experimental setting and report performance using the Avg@4 and Pass@4 metrics.

\noindent \textbf{LLMs and Datasets.} We source the training data from Ultra-FineWeb~\citep{wang2025ultra} and employ \texttt{GLM-4.6} to generate 10K synthetic trajectories. These trajectories are then used to fine-tune both \texttt{Qwen3-8B} and \texttt{Qwen3-32B}. Additionally, we leverage the same 10K samples to train a data synthesizer model based on \texttt{Qwen3-8B}.

\noindent \textbf{Baselines.} We compare our synthetic data against the following open-source datasets: (1) \textbf{APIGEN-MT}~\citep{prabhakar2025apigen}, (2) \textbf{Simia-Tau}~\citep{li2025simulating}: We sample 50K multi-turn data from this dataset, (3) \textbf{MUA}~\citep{zhao2025mua}, and (4) \textbf{TOUCAN}~\citep{xu2025toucan}: We sample 50K multi-turn data from this dataset. Notably, APIGEN-MT and Simia are in-domain training data generated in the $\tau$-bench environment (Airline and Retail).

\noindent \textbf{Training Details.} For supervised fine-tuning, we set the learning rate to $5\times10^{-6}$ and train for two epochs. Fine-tuning is performed using LLaMA-Factory~\citep{zheng2024llamafactory} under a full-parameter setting. Further details are provided in Appendix~\ref{app:hyper}.

\subsection{Performance of GEM Synthesis Pipeline}
\paragraph{BFCL V3 Results}
Table~\ref{tab:bfcl_results} presents performance comparisons on the BFCL V3 benchmark. The results of Simia are not included, as we observed its scores to be generally low, likely due to its reliance solely on in-domain $\tau$-bench data.
Our proposed GEM method demonstrates clear improvements over baseline models at both the 8B and 32B scales. In the 8B category, Qwen3-8B-GEM achieves an overall accuracy of 30.25\%, significantly surpassing the base Qwen3-8B model and outperforming other open-source baselines such as APIGEN-MT and TOUCAN.
The performance gain is even more pronounced in the 32B category, where Qwen3-32B-GEM attains an accuracy of 44.88\%. This result not only exceeds open-source synthesized datasets like APIGEN-MT and MUA by a wide margin but also outperforms proprietary large-scale models, including GPT-4.1 (38.88\%) and DeepSeek-V3.2-Exp (37.38\%).
These findings confirm that our data synthesis strategy effectively enhances function-calling capabilities across various multi-turn interaction categories.

\paragraph{$\tau^2$-Bench Results}
\begin{wrapfigure}{r}{0.4\textwidth}
    \centering
    \setlength{\tabcolsep}{2.5pt}
    \resizebox{0.9\linewidth}{!}{%
    \begin{tabular}{l cc cc}
    \toprule
    \multirow{2}{*}{\textbf{Model}} & \multicolumn{2}{c}{\textbf{Airline}} & \multicolumn{2}{c}{\textbf{Retail}} \\
    \cmidrule(lr){2-3} \cmidrule(lr){4-5} 
     & \textbf{Avg@4} & \textbf{Pass@4} & \textbf{Avg@4} & \textbf{Pass@4} \\
    \midrule
    \multicolumn{5}{c}{\textit{8B Models}} \\
    \midrule
    Qwen3-8B & 13.00 & 18.00 & 38.16 & 66.67 \\
    APIGEN-MT & 23.50 & 42.00 & 42.54 & 69.30 \\
    Simia & \textbf{35.50} & \textbf{52.00} & 43.20 & 70.18 \\
    TOUCAN & 20.50 & 42.00 & 26.32 & 52.51 \\
    MUA & 20.50 & 40.00 & 30.26 & 61.40 \\
    \rowcolor{gemblue} Qwen3-8B-GEM & 22.00 & 40.00 & \textbf{44.52} & \textbf{75.44} \\
    \midrule
    \multicolumn{5}{c}{\textit{32B Models}} \\
    \midrule
    Qwen3-32B & 21.00 & 40.00 & 43.20 & 70.18 \\
    APIGEN-MT & 36.00 & 52.00 & 44.52 & 74.56 \\
    Simia & \textbf{38.00} & \textbf{62.00} & 48.03 & 73.68 \\
    TOUCAN & 30.00 & 52.00 & 43.86 & 72.81 \\
    MUA & 33.00 & 54.00 & 49.56 & 80.70 \\
    \rowcolor{gemblue} Qwen3-32B-GEM & 35.50 & 56.00 & \textbf{55.48} & \textbf{86.84} \\
    \bottomrule
    \end{tabular}%
    }
    \caption{\textbf{Results on $\tau^2$-bench.} We report avg@4 and pass@4 metrics for Airline and Retail domains.}
    \label{tab:tau2_results}
\end{wrapfigure}
Table~\ref{tab:tau2_results} presents the performance on the $\tau^2$-bench benchmark, covering the Airline and Retail domains. We would like to emphasize that our models are trained on synthetically generated data that is strictly \textit{out-of-domain} with respect to the $\tau^2$-bench test sets, while APIGEN-MT and SIMIA are synthesized datasets within the $\tau$ environment (Retail and Airline), which can be regarded as in-domain training data.
Despite this, our approach still demonstrates comparable performance.
At the 8B scale, Qwen3-8B-GEM remains highly competitive, achieving results comparable to models like SIMIA that are fine-tuned on in-domain synthetic data, and surpassing APIGEN-MT in the Retail domain with a Pass@4 score of 75.44\% versus 69.30\%.
At the 32B scale, our model exhibits strong generalization, outperforming both SIMIA and MUA in the Retail domain with a Pass@4 of 86.84\% and delivering competitive results in the Airline domain. 
This indicates that our text-based synthesis pipeline instills a fundamental understanding of tool-use reasoning that transfers effectively to unseen, real-world domains, matching or even exceeding the performance of models trained on more domain-aligned data distributions.
Overall, the results underscore the effectiveness of our proposed paradigm.

\subsection{Performance of GEM Synthsizer}
We use 10K trajectories generated by GEM synthesizer to fine-tune \texttt{Qwen3-8B}, as shown in Table~\ref{tab:synthesizer}.
When using the same data source (Ultrafineweb), our GEM-Synthesizer achieves an overall accuracy of 28.38\% on the BFCL dataset, which is close to the performance obtained by directly using synthesized data generated by \texttt{GLM-4.6}, and also yields the Pass@4 score (73.68\%) in the $\tau^2$ Retail domain.
We also use another textual source, Wikihow~\citep{koupaee2018wikihow}, to show the  generalization of our synthesizer. The \textit{GEM-Synthesizer} also achieves a high overall performance on both BFCL (28.50\%) and on the $\tau^2$-bench Airline domain. These findings indicate that our synthesizer can generate high-quality tool-use trajectories in a low-cost, end-to-end way, thereby enabling stronger generalization in complex tool-use scenarios.

\begin{table*}[t]
\centering
\resizebox{\linewidth}{!}{%
\begin{tabular}{l ccccc cccc}
\toprule
\multirow{3}{*}{\textbf{Model}} & \multicolumn{5}{c}{\textbf{BFCL V3 Multi-Turn}} & \multicolumn{4}{c}{\textbf{Tau2}} \\
\cmidrule(lr){2-6} \cmidrule(lr){7-10}
 & \textbf{Overall} & \textbf{Base} & \textbf{Miss} & \textbf{Miss} & \textbf{Long} & \multicolumn{2}{c}{\textbf{Airline}} & \multicolumn{2}{c}{\textbf{Retail}} \\
 \cmidrule(lr){7-8} \cmidrule(lr){9-10}
 & \textbf{Acc} & \textbf{Acc} & \textbf{Func} & \textbf{Param} & \textbf{Context} & \textbf{Avg@4} & \textbf{Pass@4} & \textbf{Avg@4} & \textbf{Pass@4} \\
\midrule
Qwen3-8B (Base) & 18.00 & 24.00 & 17.00 & 13.50 & 17.50 & 13.00 & 18.00 & 38.16 & 66.67 \\
\midrule
+ GEM-GLM (Ultrafineweb) & 30.25 & 40.00 & 30.00 & 28.00 & 23.00 & 22.00 & 40.00 & 44.52 & 75.44 \\
+ GEM-Synthesizer (Ultrafineweb) & 28.38 & 41.50 & 23.50 & 27.50 & 21.00 & 26.00 & 40.00 & 42.11 & 73.68 \\
+ GEM-Synthesizer (Wikihow) & 28.50 & 43.50 & 22.00 & 27.00 & 21.50 & 25.00 & 42.00 & 39.67 & 68.42 \\
\bottomrule
\end{tabular}%
}
\caption{\textbf{Effects of data synthesizer and text sources.} We compare the impact of different trajectory generation models (GLM vs. Trained Synthesizer) and data sources (Ultrafineweb and Wikihow) on the Qwen3-8B base model. The best performance is highlighted in \textbf{bold}.}
\label{tab:synthesizer}
\end{table*}
\subsection{Ablation Study}
\label{sec:ablation}
\begin{wrapfigure}{r}{0.6\textwidth}
    \centering
    \setlength{\tabcolsep}{3pt}
    \resizebox{0.95\linewidth}{!}{%
    \begin{tabular}{l ccccc}
    \toprule
    \multirow{2}{*}{\textbf{Model}} & \textbf{Overall} & \textbf{Multi} & \textbf{Miss} & \textbf{Miss} & \textbf{Long} \\
     & \textbf{Acc} & \textbf{Base} & \textbf{Func} & \textbf{Param} & \textbf{Cxt.} \\
    \midrule
    Qwen3-8B (Base) & 18.00 & 24.00 & 17.00 & 13.50 & 17.50 \\
    \rowcolor{gemblue} Qwen3-8B-GEM & 30.25 & 40.00 & 30.00 & 28.00 & 23.00 \\
    \textit{(w/o Refine)} & 26.00 & 33.50 & 23.50 & 27.00 & 20.00 \\
    \textit{(w/o LLM-Based Check)} & 27.38 & 37.50 & 25.00 & 24.00 & 23.00 \\
    \midrule
    Qwen3-32B (Base) & 28.35 & 44.00 & 24.50 & 25.50 & 29.50 \\
    \rowcolor{gemblue} Qwen3-32B-GEM & 44.88 & 52.00 & 40.00 & 38.50 & 49.00 \\
    \textit{(w/o Refine)} & 32.50 & 40.00 & 30.00 & 27.00 & 33.00 \\
    \textit{(w/o LLM-Based Check)} & 44.25 & 52.00 & 40.50 & 40.00 & 44.50 \\
    \bottomrule
    \end{tabular}%
    }
    \caption{\textbf{Ablation study.} We conduct ablation study with BFCL V3 benchmark.}
    \label{tab:ablation}
\end{wrapfigure}


To validate the effectiveness of our synthetic data pipeline, we conduct an ablation study with BFCL V3 benchmark focusing on two key components: the refinement stage and the LLM-based check as shown in Table~\ref{tab:ablation}. We report the ablation results on $\tau^2$-bench in Appendix~\ref{app:ablation_tau}.

\noindent\textbf{Effect of Refinement.} The refinement stage leads to a substantial performance improvement. For example, it raises the overall accuracy of Qwen3-32B from 32.50\% to 44.88\%, a gain of over 12 percentage points. This process increases the complexity and quality of the synthetic trajectories, which in turn enables more effective learning of multi-turn tool use. We report more details in Appendix~\ref{app:refinement} to illustrate out refinement strategy significantly enhances the overall complexity of the synthesized data.
Notably, even the original trajectories extracted directly from the original text (though relatively simpler) still provide valuable training signals and contribute to improved tool-calling capability. This indicates that more effectively leveraging information from the original text to synthesize high-quality tool-calling trajectories is a promising research direction. 

\noindent\textbf{Effect of LLM-Based Check.} This stage consistently improves results by filtering out samples with hallucinations or inconsistencies. For the 8B model, it raises overall accuracy from 27.38\% to 30.25\%. This mechanism also contributes to stronger performance across both model scales.


\subsection{Data Analysis}
\begin{figure*}[t]
    \centering
    \begin{subfigure}[t]{0.33\textwidth}
        \centering
        \includegraphics[width=\linewidth]{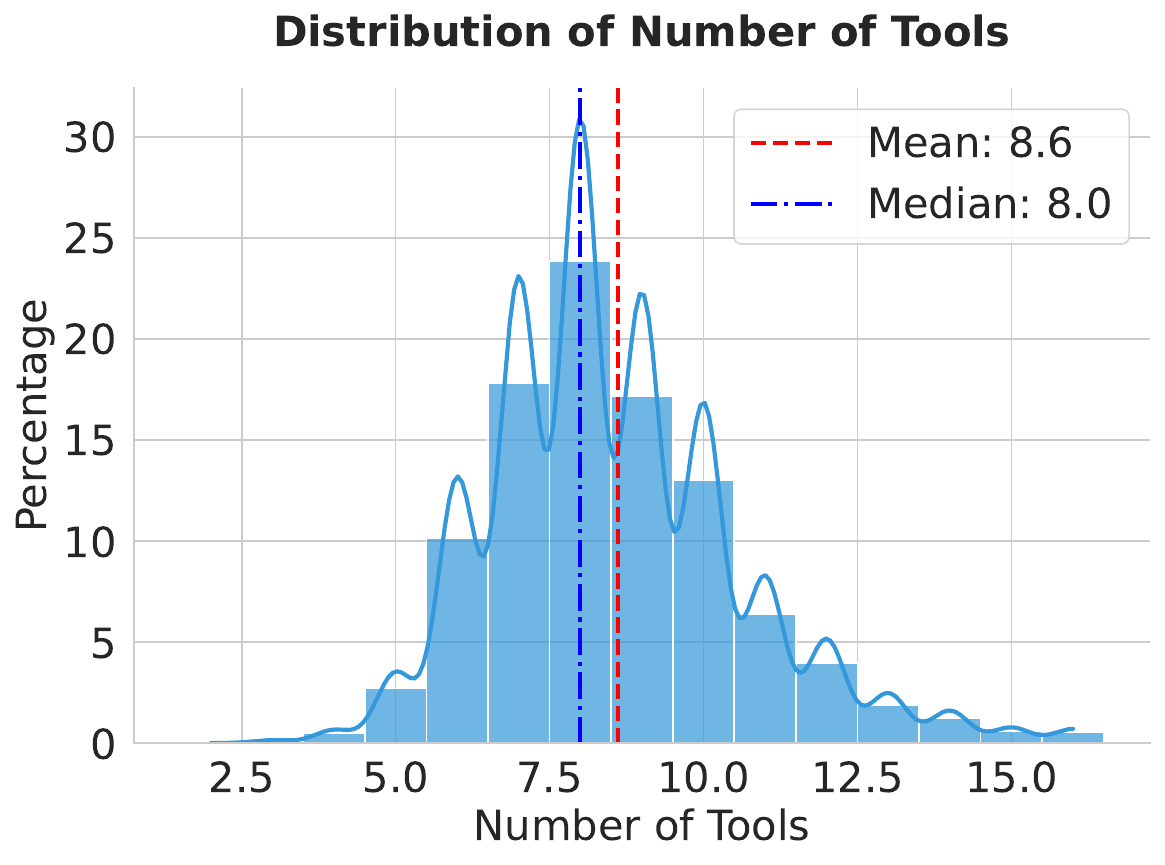}
        \label{fig:analyze_tools}
    \end{subfigure}\hfill
    \begin{subfigure}[t]{0.33\textwidth}
        \centering
        \includegraphics[width=\linewidth]{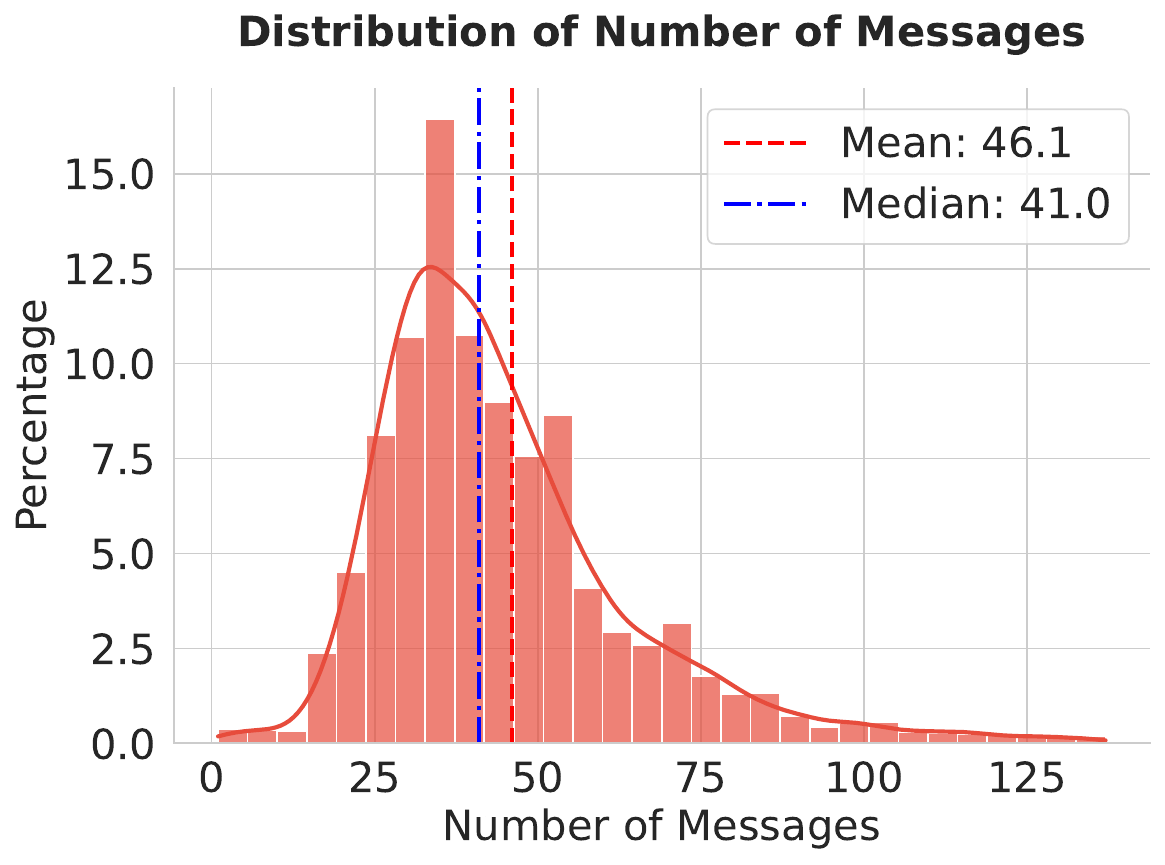}
        \label{fig:analyze_messages}
    \end{subfigure}
    \begin{subfigure}[t]{0.33\textwidth}
        \centering
        \includegraphics[width=\linewidth]{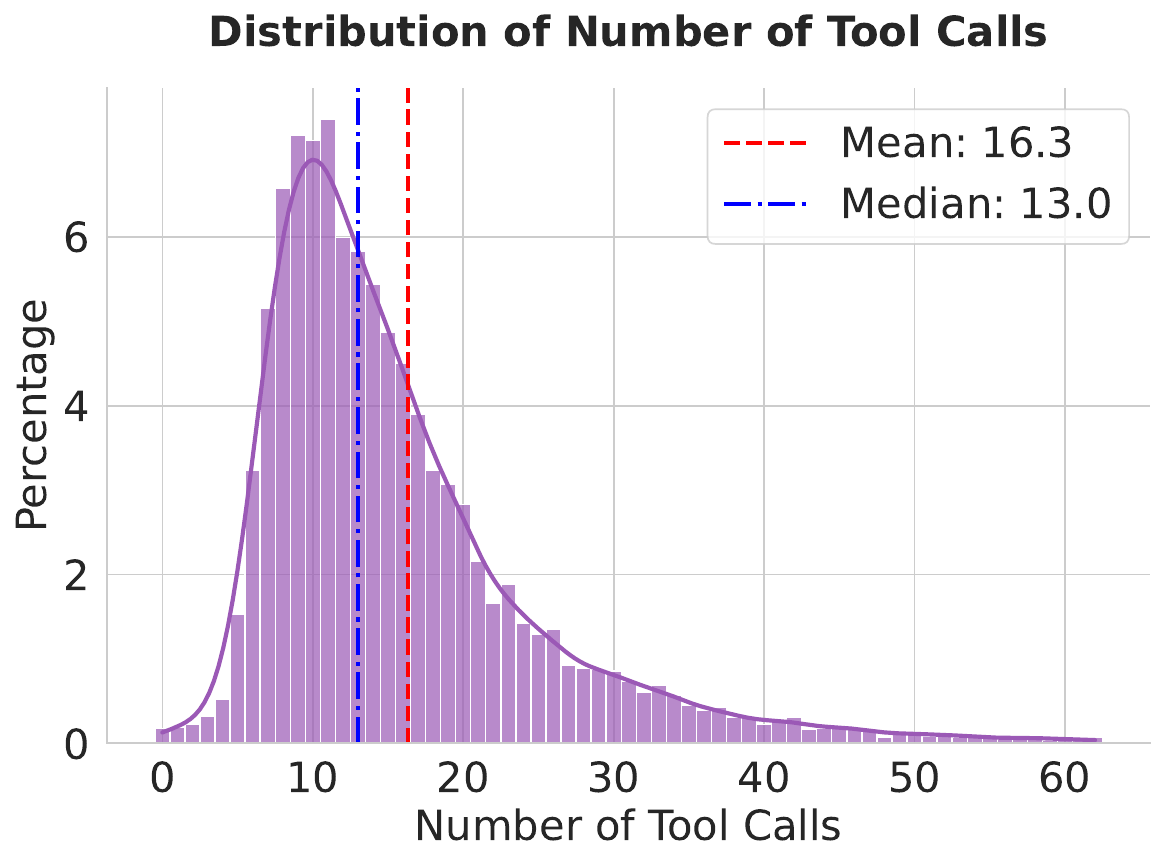}
        \label{fig:analyze_messages}
    \end{subfigure}
    \vspace{-1em}
    \caption{Data analysis. We find the synthesized trajectories are overall complex.  \textbf{Left:} distribution of number of tools. \textbf{Mid:} distribution of number of messages. \textbf{Right:} distribution of number of tool calls in each trajectory.}
    \label{fig:dataset_analysis}
\end{figure*}

Figure~\ref{fig:dataset_analysis} presents a statistical overview of our synthesized multi-turn tool-use trajectories, highlighting three key dimensions: the number of distinct tools employed, the number of messages per dialogue, and the total number of tool calls.
On average, each trajectory involves 8.6 distinct tools, indicating that the synthesis process requires models to meaningfully select and combine multiple tools within a single task.
Moreover, the trajectory contains an average of 46 turns. The considerable length of these dialogues ensures that tasks cannot be resolved simply, helping models learn to maintain context, track task state, and engage in multi‑step planning over prolonged dialogues. In comparison, existing open‑source datasets such as APIGEN-MT~\citep{prabhakar2025apigen} average around 18.5 turns, while TOUCAN~\citep{xu2025toucan} contains only about 6.24 turns. 
Finally, each trajectory averages 16.3 tool calls. while APIGEN-MT contains only an average of 4.3 tool calls. This high frequency underscores the multi‑step, tool‑driven nature of the synthesized tasks.
Taken together, these statistics demonstrate that our synthesized dataset exhibits substantial diversity and interaction depth, making it well‑suited for training models in multi‑turn tool‑use scenarios.


\subsection{Case Study}
\begin{figure*}[]
    \centering
    \includegraphics[width=\linewidth]{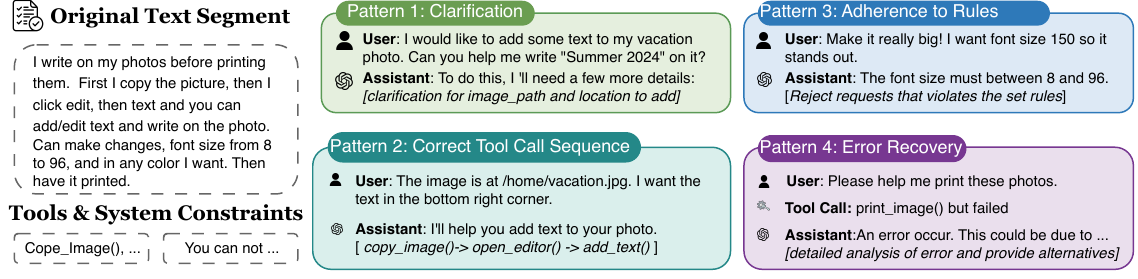}
    \caption{Case study of our generated multi-turn tool-use trajecory.}
    \label{fig:case_study}
\end{figure*}
For the case study illustrated in Figure~\ref{fig:case_study}, this synthesized trajectory is based on a real-world photo-editing scenario. 
The synthesis process first extract diverse tools from the text description, and identifies key constraints from the user's actions, such as the restriction that font size must be between 8 and 96 points. 
Building on these tools and rules, the dialogue encompasses a variety of realistic interaction patterns: proactively clarifying missing parameters (i.e., image path and text placement), invoking tools in the correct sequence, rejecting requests that violate constraints (i.e., exceeding the font size limit) while offering compliant alternatives, and recovering from errors by retrying with another available printer. 
It guides the model to learn how to validate inputs, adhere to system constraints, execute tasks step by step, and respond flexibly to errors within multi-turn interactions, thereby achieving reliable tool-use capabilities.
We also show a complete synthesized trajectory in Appendeix~\ref{app:full_case} to help readers better understand our methodology.
\section{Related Work}

\paragraph{Tool-Use Data Synthesis} To equip LLM-based agents with tool-calling capabilities, prior works focus on synthesizing tool-use training data. 
ToolBench~\citep{qin2023toolllm} builds a large-scale function-calling dataset over thousands of APIs. 
ToolACE~\citep{liu2024toolace} designs an automatic agentic pipeline that iteratively expands an API pool and synthesizes verified function-calling traces with high complexity and diversity. 
For multi-turn interaction, APIGen-MT~\citep{prabhakar2025apigen} generates structured task blueprints with ground-truth action sequences and then simulates realistic human–agent dialogues grounded in executable APIs. MagNet~\citep{yin2025magnet} represents multi-tool workflows as graph-structured function paths and converts them into multi-turn conversations with executable calls. ToolACE-MT~\citep{zeng2025toolace} adopts a non-autoregressive framework that drafts entire dialogues in one shot and refines them via iterative editing and verification.
TOUCAN~\citep{xu2025toucan} crawls MCP servers and synthesizing 1.5M of tool-use data. 
Unlike prior works that rely on pre-defined tools, our work introduces a novel paradigm that directly extracts multi-turn trajectories from text, thereby unlocking a authentic and scalable source of tool-use agentic data.

\paragraph{Tool-Use Capability Evaluation}
To systematically assess tool-use capabilities of LLM agents, a series of benchmarks have been proposed.
ToolBench~\citep{qin2023toolllm} focuses on evaluating whether models can translate natural-language instructions into correct API calls across thousands of real-world tools.
The Berkeley Function Calling Leaderboard (BFCL)~\citep{patilberkeley} provides a large-scale, syntax- and semantics-aware evaluation of function calling across diverse domains and programming languages, using AST-based checks to measure the accuracy and compositionality (e.g., parallel calls) of model-generated function invocations.
ACEBench~\citep{chen2025acebench} evaluates multi-turn tool-use capability from three perspectives: normal, special, and agent.
$\tau$-bench~\citep{yao2024tau} emulates dynamic conversations between a simulated user and a tool-augmented agent in domain-specific scenarios, jointly evaluating task success, tool selection, and adherence to domain policies. $\tau^2$-bench~\citep{barres2025tau} further extends this line by introducing dual-control environments where both the agent and the user can invoke tools, highlighting the challenges of coordinating tool use and guiding user actions in more realistic customer-service-style tasks. VitaBench~\citep{he2025vitabench} focuses on evaluating agent's performance on life-serving simulation environment.

\section{Conclusion}
This paper presents a novel paradigm and the agentic data synthesis pipeline, which directly synthesize multi-turn tool-use trajectories from text corpora, effectively bypassing the dependency on predefined tools.
Models trained on our data achieve significant performance gains on benchmarks, demonstrating the potential of leveraging open-world textual knowledge as a scalable source for advancing autonomous agents.

\bibliographystyle{unsrtnat}
\bibliography{ref}

\clearpage
\section*{Appendix}
\appendix
\section{Prompts}
\label{app:prompt}
\subsection{Tag Annotation}
\label{app:metadata}
\lstset{
    basicstyle=\ttfamily\small,
    breaklines=true,  
    backgroundcolor=\color{gray!5},
    frame=none,
    columns=fullflexible,
    keepspaces=true,
    showstringspaces=false
}

\begin{tcolorbox}[
    colback=gray!5!white, 
    colframe=gray!75!black, 
    title={Domain \& Category Annotation}, 
    fonttitle=\bfseries, 
    sharp corners,
    width=\textwidth
]
\begin{lstlisting}
Determine whether the following text contains multi-step operations involving the use of an APP, website, computer, or other machine (such as robot, elevator, etc), if contains, generate one sentence summary of the task and identify the platform, domain and task category of the multi-step task.

# Detaild Instruction
1. Platform category: operator, computer, phone, machine, other
2. Domain category: adult, arts_and_entertainment, autos_and_vehicles, beauty_and_fitness, books_and_literature, business_and_industrial, computers_and_electronics, finance, food_and_drink, games, health, hobbies_and_leisure, home_and_garden, internet_and_telecom, jobs_and_education, law_and_government, news, online_communities, people_and_society, pets_and_animals, real_estate, science, sensitive_subjects, shopping, sports, travel_and_transportation
3. Task category: databases, multimedia_processing, cloud_platforms, calendar_management, cryptocurrency, location_services, communication, search, file_systems, web_scraping, ecommerce_and_retail, customer_data_platforms, developer_tools, virtualization, version_control, research_and_data, aigc, travel_and_transportation, note_taking, language_translation, rag_systems, security_and_iam, social_media, monitoring, weather_services, customer_support, blockchain, knowledge_and_memory, financial_trading, marketing, enterprise_business_intelligence, transportation_logistics, iphone_android, smart_home, education_elearning, robot_control, website_control, gaming_entertainment

# Input
[text]

# Output Format
<multi_step>False</multi_step>
Or:
<multi_step>True</multi_step>
<summary>xxx</summary>
<domain>Shopping, Sports</domain>
<platform>Operator</platform>
<task>customer_support</task>
\end{lstlisting}
\end{tcolorbox}

\subsection{Workflow \& Tool Discovery}

\begin{tcolorbox}[
    colback=gray!5!white, 
    colframe=gray!75!black, 
    title={Workflow \& Tool Discovery}, 
    fonttitle=\bfseries, 
    sharp corners,
    width=\textwidth,
    breakable,  
    enhanced jigsaw  
]
\begin{lstlisting}
You are an program design expert. 
Given a workflow description in a scenario, your task is to design multiple functions to translate the execution process of this workflow into program.

# Instruction
1. Extract all intermediate steps in the workflow, if the text contains multiple workflows, output them in a list.
2. Convert **every** step to a function and represent them as an execution graph (i.e. (login)->(search_query)->..)
3. Based on the execution graph, generate real API calls that populate the tools with reasonable parameters, simulating a use case of actual tool invocation steps.
4. Provide detailed API definitions used in the above process.
5. Follow the following steps to generate more complex workflows and tools:
- Workflow Exploration: You need to explore multiple workflows or complex constraints that may exist in the document
    - These workflows represent the possible interaction patterns of a real user-agent.
    - Dependencies: "X must happen before Y".
    - Uniqueness/Limits: "Only one Admin allowed", "Name must be unique".
    - Conditionals: "If user is X, they cannot do Y".
- Tool Design (Functional API Level)
    Design a set of JSON-schema tools based on the text.
    - The required parameters of a tool need to be carefully considered and designed, mirroring the logic of the real world. For example, viewing system data typically requires authorization authentication, and providing user ID, product ID, etc.
    - It mimics a database structure and provides read and write tools. For example, it provides tools for querying user information, along with corresponding tools for modifying user information.
    - Each tool's name should be short and readable, semantically clear and general, reusable (e.g., "flight_search" rather than "flight_detailed_search_for_tom_2025")
    - Each tool should implement a single, coherent capability. It should not bundle multiple unrelated or multi-stage workflows into one tool. (e.g., create two tools "plan_trip" + "book_trip" rather than only one tool named "plan_and_book_trip")
    - Each tool's parameters should be explicitly defined in the schema with clear types and meanings. Parameter names should be self-explanatory rather than cryptic (e.g., use "check_in_date" with type "string" and a short description, rather than a vague parameter named "d1").
    - The majority of tools describe functional data operations that either retrieve information from or modify the state of the environment (e.g., get_status, update_permissions)

# Workflow Description
[text]

# Output Format
<workflow>
<description>short task description</description>
<steps>Step1: ...\nStep2: ...</steps>
<execution_graph>(api_name1)->(api_name2, api_name3)->..</execution_graph>
<actions>[{"name":"api_name", "arguments": {"arg_name": "value", ...}}, ... (more API calls as required)]</actions>
<tools>[{"name":"api_name1","description":"","inputSchema":{"type":"object","properties":{"arg_name1":{"type":"","description":""},"arg_name2":{"type":"","description":""}},"required":["arg_name2"]}},{"name":"api_name2","description":"","inputSchema":{"type":"object","properties":{"arg_name1":{"type":"","description":""},"arg_name2":{"type":"","description":""}},"required":["arg_name2"]}}]</tools>
</workflow>

<workflow>
(more workflows)
</workflow>
\end{lstlisting}
\end{tcolorbox}

\subsection{Trajectory Generation}

\begin{tcolorbox}[
    colback=gray!5!white, 
    colframe=gray!75!black, 
    title={Trajectory Generation}, 
    fonttitle=\bfseries, 
    sharp corners,
    width=\textwidth,
    breakable,  
    enhanced jigsaw  
]
\begin{lstlisting}
You are tasked with generating high-quality multi-turn dialogue trajectories based on a given text document. The trajectory should demonstrate an AI assistant helping users complete tasks while strictly following domain-specific rules and constraints.

You will be provided with:
- A list of Available Tool Candidates;
- A source text document that contains the description of the scenario and task steps;

## Completion Requirements 
1. System Prompt: Extract and explicitly state ALL important domain-specific rules and constraints from the source text document.
Example:
<system>
You are a agent specilized in retail domian. Here are some basic rules to follow:
- An order can only be cancelled if its status is 'pending' ...
- Modify action can only be called once, and will change the order status to 'pending (items modifed)' ...
- ...
</system>

2. User Task: Create natural, progressive user requests that test the system's rule enforcement and constraint handling. Here are some features:
- Naturalness: Requests should reflect real-world use cases
- Ambiguity: User requests are often incomplete, requiring the assistant to analyze or clarify them.
  Example: <user> I want to cancel order #W2575533. </user> (the user do not provide the specific reason, and the assistant should ask for clarification); <user> Recommend me a desktop. I often go out. </user> (the user do not explicitly state the attribute of the item, but the assistant should analyze and know it based on the stated preference)
- Consistent: User's intention, persona, and their behavior should be consistent across the dialogue.
- Complex: The user's request is challenging enough to test the assistant's ability. Users can make requests that violate domain rules and are not allowed to alert the assistant (e.g., do not ask the model to verify the order status first).
  At least in one turn, the user's request is very complex and require assistant to handle it carefully.
  Example 1: 
  <user> I need to make several changes to my order #W2575533. Can I change the E-Reader to a different size, swap the Garden Hose color, and also update my shipping address </user>
  (Requires assistant to: check order status, verify each item can be modified, handle address change separately, remind about one-time modification limit)
  Example 2:
  <user> Check my tire pressures. If any of them are low, find me the nearest service station and also check if I have enough fuel to get there
  (Requires: check tire pressure, evaluate condition, conditionally call find_nearest_shop, check fuel level, calculate if sufficient)
  Example 3:
  <user> I'm planning a three-day trip starting from Hangzhou, and I need help creating an itinerary. One more thing about the second day - I'm trying to be smart about my budget. If I end up booking a luxury hotel that costs 800 CNY or more per night, then I need to be more careful with other expenses: my total spending on both restaurants (lunch and dinner) should stay under 350 CNY, both restaurants should be rated at least 4.0 stars,and the afternoon attraction ticket needs to be less than 120 CNY. </user>
  (Requires: check multiple constraints)

3. Assistant: Generate Responses that Demonstrate Rule Enforcement, Clear Communication, and Intelligent Problem-Solving:
- Reasoning and Adaptive Planning: The Assistant should reason through problem contexts and plan appropriate steps. Sometimes users may not be able to directly provide the parameters for tool calls, and the assistant needs to accurately consider whether the parameter values can be obtained through other information and tools.
- Precondition Checks: Before executing tasks, the Assistant should validate any necessary preconditions (e.g., authenticating identity, verifying the status of an order)
- Domain Rules and Constraints: The Assistant must follow domain-specific rules at all times. Ensure the assistant' tool call and response genuinely addresses those requirements.
- User-Centric Principle: The Assistant must accurately understand and satisfy all user needs and preferences without breaking domain rules.  For example, if user states "prefers A, wants B, and tell me C", the assistant should satisfy all requirements.
- No Hallucination: 
  - Context Faithfulness: Maintain absolute fidelity to tool outputs. If data contradicts user expectations, explicitly report the discrepancy instead of distorting facts to force a match.
  - Tool call arguments: The Assistant must only use argument values that are explicitly provided or implied by the user. It must not fabricate IDs, names, or other parameters; if any required value is missing or unclear, the Assistant should ask the user to supply it before calling the tool.
- Consequence-aware: Before executing any write operation that modifies the environment, the Assistant must actively think and assess its impact. For changes that are irreversible, the assistant should obtain explicit user comformation before proceeding.
- Limitations: If the current request is beyond the assistant's ability, the Assistant must communicate this limitation clearly. For example: If the user's needs are beyond domain rules, the Assistant must explain the limitation but should still respect the user's needs. (e.g., "I cannot do A, but I can do B. Should I proceed with B?" -> Wait for conformation).
- Correct Tool Calls:
  - Tool call format: `<func>{"name":"exact_tool_name", "arguments": {"arg": "value"}}</func>`.
  - Exact Matching: The Assistant must ensure that tool calls exactly match the available candidate tools. Calling tools that do not exist is prohibited.
  - Parameter Validation: The arguments passed to the tools must exactly match the tool definitions, including the required parameters. The Assistant should avoid any missing parameters and validate that the values are accurate.

4. Tool Responses: Structured and Complete
- Correctness of Tool Call: A success response should only be returned if the tool call is correct (including both the correct function name and parameters). If any part of the tool call is incorrect or incomplete, the tool should return a failure response, indicating what went wron
- Success Response: The tool's success response must return all relevant information in a well-structured format (e.g., JSON). This includes not just the requested data, but also any other relevant details, such as order ID, status, product ID, item ID, etc., in case the user requires further context.
- Error Response: When there is an error, the tool should return only the error message. No additional information should be provided that directly aids the Assistant in recovering from the error.
- Consistent Response Structure: The format and content of the tool's responses should remain consistent throughout the conversation. This ensures clarity and reliability in the tool's output, helping to maintain a smooth user experience.
- Don't confuse the order between turns: 
  - a user message or tool result should be followed by an assistant message. (<user>...<user> / <tool>...</tool> -> <assistant> ... </assistant>)
  - If the assistant message includes a tool call, it can be followed by a tool message; (<assistant>...<func>...</func></assistant> -> <tool>...</tool>)
  - otherwise, it should be followed by a user message. (<assistant> ... </assistant> -> <user> ... </user>)
  - Tool result cannot followed by user message. (MUST NOT OUTPUT: <tool>...</tool> -> <user>...</user> (incorrect example))

5. Trajectory Pattern: The trajectory should exhibit multiple interaction patterns, with at least 3 distinct patterns appearing in total. Each individual pattern should be used at most 2 times within a single trajectory.
[Pattern 1: Domain Rules & User Need conflicts]
[Pattern 2: Error Recovery]
[Pattern 3: Clarification and Disambiguation]
[Pattern 4: Assistant's Multi-hop Reasoning]
[Pattern 5: Assistant's Awareness of Domain Rules]

## Other Requirements 
- Always use English.
- The whole trajectory should be reasonable, realistic (conform to real-world dialogue scenarios and interactions), and fit the context of multi-turn tool usage. 
- Don't confuse the order between rounds: a user message or tool result should be followed by an assistant message. If the assistant message includes a tool call, it can be followed by a tool message; otherwise, it should be followed by a user message. Tool result cannot followed by user message.
- Tool call format: 
<func>
{{"name": "exact_tool_name_in_toolsets", "arguments": {{"arg": "value"}}}}
</func>

## Given Inputs 
### Available Tool Candidates
{candidate_tools}

### A source text document consisting of Task Steps Description
{current_task}

## Output Format

You must STRICTLY follow the following output format.
Ensure ALL tags are properly opened and closed. Conversations like "<tool> ... </assistant>", "<assistant>...</user>" are wrong!!!

<system>
...
</system>
<user>
...
</user>
<assistant>
...
<func>
{{"name": "...", "arguments": {{...}}}}
</func>
</assistant>
<tool>
...
</tool>
\end{lstlisting}
\end{tcolorbox}

\subsection{Trajectory Refinement}

\begin{tcolorbox}[
    colback=gray!5!white, 
    colframe=gray!75!black, 
    title={Trajectory Refinement}, 
    fonttitle=\bfseries, 
    sharp corners,
    width=\textwidth,
    breakable,  
    enhanced jigsaw  
]
\begin{lstlisting}
You are rewriting a complex, realistic multi-turn tool-use trajectory for agenic training.
Goal: The trajectory should be complex, natural, no-hallucination and show the wisdom of the assistant.
It must show the assistant's ability for correct tool use, reasoning, context understanding, and communication skills with users.
Please carefully consider what problems exist with the existing synthetic trajectories and how to improve them to create high-quality trajectories.
You can follow the following guidelines.

# Guideline

### System Prompt Complexity
You need to refine and upgrade the constraints of the system prompts to make them more structured, systematic, and consistent with real-world logic, in order to fully test the agent's ability to make correct tool calls in complex scenarios. You also need to define the database schema in the system prompts, and the data structure returned by the tool will be based on this.

### User Request Complextity & Naturalness
- Natural: Natural requests may include colloquial language, implied context, vague references to prior steps, or real-world motivations (e.g., saving time/money, lose weight). Avoid overly formal or purely instructional language.
- User diversity: Create a user profile and maintain the user's personality and characteristics throughout the conversation history.
- Requires Deep Analysis & Reasoning: The user's request must necessitate careful analysis and multi-step reasoning to identify the correct tool(s) and determine appropriate parameter values.
- Requires Analysis of Tool Dependencies & Outputs: The request should force the assistant to understand dependencies between tools and use outputs from previous steps to decide which tool to invoke next.
- In at least 1 turn, the user's request contains multiple constraints, including explicit constraints, implicit requirements that require the assistant to infer.
- In at least 1 turn, The user asks a question that can only be answered by reasoning across outputs from multiple tool calls in the long context.
- Challenging Pitfalls: **MUST INCLUDE AT LEAST 1-2 PITTFALLs** for one trajectory.
  This pattern sets traps to test the assistant's ability to correctly make robust tool calls based on rules, constraints or user preferences. 
  The trajectory must include request for this kind of challenging tool calls, and the assistant must explicitly analyze and identify these pitfalls to achieve robust tool calls.

### Assistant Intelligibility
Demonstrate the following capabilities of the assistant:

(1) Communication Skills
- User Intent Understanding: Accurately interpret the underlying goals and context of the user's request.
- Confirmation & Clarification: Proactively confirm key details or ask the user to clarify ambiguous information when necessary.
- Capability Limitation Awareness: Clearly communicate the boundaries and limitations of the assistant's available capabilities.
- Result Explanation & Summarization: Interpret, distill, and present tool outputs or complex information in a structured, understandable manner.
- Proactive Assistance: Anticipate potential user needs based on context and offer helpful suggestions in advance.

(2) Robust Tool-Calling Capability
- Tool Selection: Choose the most appropriate tool from the available set based on task requirements.
- Sequential Tool Usage: Plan and execute multi-tool workflows with correct dependencies and order.
- Parameter Handling: Correctly construct and validate complex argument structures (e.g., nested objects, lists), respecting type, range, and format constraints.
- Result Analysis: Parse and evaluate tool responses, extract relevant information, and determine validity for subsequent steps.
- State Tracking: Maintain awareness of completed and pending steps in a multi-step task.
- Constraint Analysis: Identify and adhere to real-world constraints and business rules (e.g., date ranges, mutually exclusive fields, batch limits).
- Error Handling: Gracefully manage tool failures, diagnose error causes, and adjust strategy or inform the user appropriately.
- Context Management: Effectively retain and utilize conversation history to ensure coherence across complex interactions.

(3) Reasoning & Execution Ability
- Planning & Task Decomposition: Break down complex or vague user requests into clear, executable step-by-step plans.
- Prerequisite Management: Recognize and acquire necessary information or conditions before executing tasks (e.g., querying environment, requesting user input).
- Verification and Validation: Perform essential checks before critical operations, such as exploring available tools, confirming permissions, or validating inputs.

### Realistic & Complex Environment
IMPORTANT NOTES:
- The tool called in the trajectory MUST exist in candidate toolsets, with the parameter type and value correct.
- If it's necessary to add tools when constructing complex trajectories, you can add them in the final output tools block.
GOALS:
- This increases the difficulty of choosing tools. A diverse range of tools must be included, specifically reading and writing tools.
- Increase the difficulty of making totally correct tool calls (parameter type, value). For example, include structured inputs (list, dict, nested objects) and meaningful constraints.
- The tools should conform to the database schema as much as possible (may defined in system prompt), and mainly include read-write tools. You can add & modify tools.
- Unique: Tool call parameters should use unique database fields (such as user ID, product ID, etc.) as much as possible to mimic real logic.
- Realistic: Always avoid using empty placeholders. For example, do not return something like "5+ more results ...", "path=/example".
- Success Response: A success response should only be returned if the tool call is correct. The tool's response must return a complete data structure in a well-structured format (e.g., JSON). 
- Error Response: You are allowed to simulate non-simple errors that might occur in the real world. The tool's response must return a concise error message. Avoid directly telling the assistant how to solve the problem.

### Trajectory Diversity
- Reduce the frequency of repeatedly using certain tools to solve problems, avoiding the reduction of trajectory diversity, and retain only the most valuable trajectories for learning.
- Include but not limited to the following pattern, and make the following pattern more difficult, diverse and natural.
    - [Pattern 1: Environment Complexity] 
        - [1.1: Error Recovery] In multi-turn function calling, models may encounter errors, such as invalid input or failed execution that require recovery. If you think of any suitable, non-trivial, real-world scenario errors, please include this pattern.
        - [1.2: Long Context] Introduce large volumes of extraneous data to test how well the model can extract crucial details from an overwhelming array of information. 
    - [Pattern 2: Clarification] Tests the model's ability to recognize when essential information is missing from the user request. 
        - [2.1: Can be inferred from the system] The assistant actively explore ways to find the essential information and complete the task
        - [2.2: Can not be inferred from the system] The assistant actively clarifies the situation.
    - [Pattern 3: Identify Limitation] Requires the model to identify that no available function can fulfill the user request.
    - [Pattern 4: Assistant guides user operation] For example, if a user reports no internet access, the assistant uses tool calls to discover that the SIM card is not inserted, and then guides the user to insert it (this process cannot be performed by the assistant alone due to real-world phycial constraints and requires active communication and guidance between the assistant and the user).

# Input Data
Toolset: {tools}
Trajectory: {our_traj}

# Output Requirement
- Reduce redundancy: Reduce the frequency of repeatedly using certain tools to solve problems, avoiding the reduction of trajectory diversity, and retain only the most valuable trajectories for learning.
- Preserve turn order.
    - The assistant can only call the tool once per round. 
    - Each tool message must be followed by an assistant message.
    - If an assistant message contains a tool call, it must be followed by a tool message (tool result).
    - Tool messages must not be followed directly by user messages.
- If the original trajectory violates these rules or misuses tools, fix it in the rewritten version.
- Output all candidate tools (the original tools + new tools if needed)
- You must strictly follow the following output format.

# Output format
<toolsets>
All candidate tools in JSON, OPENAI format.
[
    {{
        "name": "",
        "description": "",
        "inputSchema": {{
            "type": "",
            "properties": {{ }},
            "required": []
        }}
    }},
    ...
]
</toolsets>

<system>
[role and domain rules here]
</system>
<user>
...
</user>
<assistant>
...
<func>
{{"name": "...", "arguments": {{...}}}}
</func>
</assistant>
<tool>
[concrete tool response in JSON format if tool calls are made]
</tool>
<assistant>
...
</assistant>
...
(more conversations here)
"""

\end{lstlisting}
\end{tcolorbox}

\subsection{Hallucination Detection}

\begin{tcolorbox}[
    colback=gray!5!white, 
    colframe=gray!75!black, 
    title={Hallucination Detection Prompt}, 
    fonttitle=\bfseries, 
    sharp corners,
    width=\textwidth,
    breakable,  
    enhanced jigsaw  
]
\begin{lstlisting}
You are given a multi-turn tool-use trajectory. Please evaluate the trajectory according to the following rubrics.
Your job is to score the trajectory on the following binary rubrics.
For EACH rubric, you must output 0 or 1 only, according to the criteria below.
For each rubric, if there is no hallucination of the following content throughout the entire trajectory, the corresponding rubric score is 1. 
If any single round does not meet (the condition), the corresponding rubric should be scored as 0. Be strict in your evaluation.

## Rubric

R1: Tool-call hallucination 
- Check whether any tool call uses argument values that are not provided or reasonably derivable from the dialogue context.

R2: Capability hallucination 
Check whether the assistant makes incorrect claims about what can or cannot be done.
- H2-a False inability: The user request IS solvable using the available tools, but the assistant claims it cannot be done or refuses without justification.
- H2-b Missing limitation disclosure: The user request is NOT solvable with the available tools, but the assistant proceeds as if it is solvable, or fails to clearly explain the limitation and offer the closest feasible alternative.

R3: Context hallucination 
Check whether the assistant misinterprets the ongoing context or references things that are not true in the dialogue.
- Wrongly referencing previous user constraints, preferences, or decisions.
- Cross-turn inconsistency: changing entities/values (IDs, counts, dates, constraints) without new evidence or tool output.
- Conflicting summaries: later summary contradicts earlier established facts.

## Input
{trajectory}

## Output Format
Return a single JSON object with EXACTLY these keys and integer values 0 or 1:

{{
  "R1": 0 or 1,
  "R2": 0 or 1,
  "R3": 0 or 1,
}}

Do NOT output anything else (no explanations, no comments).

\end{lstlisting}
\end{tcolorbox}

\section{Hyper-parameter Setting}
\label{app:hyper}
We use the sample hyper-parameter during the experiments. Please refer to Table~\ref{tab:hyper_sft_qwen} for details.

\begin{table}[htbp]
\centering
\resizebox{0.5\textwidth}{!}{
\begin{tabular}{l|c|l|c|}
\hline
\textbf{Hyperparams} & \textbf{Values} & \textbf{Hyperparams} & \textbf{Values} \\ \hline
learning rate        & 5e-6            & weight decay         & 0.05            \\
warmup ratio         & 0.1             & max length           & 32K            \\
lr scheduler         & cosine          & batch size           & 64              \\
epoch                & 2               & BF16                 & True            \\ 
Deepspeed            & zero3           & tool-call template   & Hermes            \\ \hline
\end{tabular}
}
\caption{SFT Hyperparameters used.}
\label{tab:hyper_sft_qwen}
\end{table}

\section{Ablation Study on $\tau^2$-bench}
\label{app:ablation_tau}

We report the ablation results of $\tau^2$-bench (Airline, Retail) in Table~\ref{tab:ablation_tau}.

\begin{table}[htbp]
\centering
\resizebox{0.5\textwidth}{!}{
\begin{tabular}{lcccc}
\hline
\multirow{2}{*}{\textbf{Model}} & \multicolumn{2}{c}{\textbf{Airline}} & \multicolumn{2}{c}{\textbf{Retail}} \\ \cline{2-5} 
                                & \textbf{Avg@4}   & \textbf{Pass@4}   & \textbf{Avg@4}   & \textbf{Pass@4}  \\ \hline
Qwen3-8B            & 13.00 & 18.00 & 38.16          & 66.67          \\
\rowcolor{gemblue} Qwen3-8B-GEM        & 22.00 & \textbf{40.00} & \textbf{44.52} & \textbf{75.44} \\
w/o refine          & \textbf{25.00} & 36.00 & 41.23          & 74.56          \\
w/o LLM-Based Check & 22.00 & 36.00 & 42.76          & 71.05          \\ \hline
Qwen3-32B           & 21.00 & 40.00 & 43.20          & 70.18          \\
\rowcolor{gemblue} Qwen3-32B-GEM       & \textbf{35.50} & \textbf{56.00} & \textbf{55.48} & \textbf{86.84} \\
w/o refine          & 31.00 & 56.00 & 40.35          & 73.68          \\
w/o LLM-Based Check & 35.00 & 52.00 & 56.80          & 82.46          \\ \hline
\end{tabular}
}
\caption{Ablation Study on $\tau^2$-bench (Airline, Retail).}
\label{tab:ablation_tau}
\end{table}

\section{More Analysis on Refinement Stage}
\label{app:refinement}
\begin{table}[htbp]
\centering
\begin{tabular}{cccc}
\hline
\textbf{Method} & \textbf{\# Number of Messages} & \textbf{\# Number of Tools} & \textbf{\# Number of Tool Calls} \\ \hline
w/o refinement  & 30.05                          & 5.01                        & 7.83                             \\
w/ refinement   & 46.1                           & 8.6                         & 16.3                             \\ \hline
\end{tabular}
\caption{\textbf{Statistics of synthetic trajectories before and after refinement.} The refinement stage significantly increases the complexity of conversation trajectories, as evidenced by the rise in the average number of messages, distinct tools used, and total tool invocations.}
\label{tab:refinement}
\end{table} 

We report the mean number of messages, tools, and tool calls per trajectory before and after the refinement stage in Table~\ref{tab:refinement}. The results demonstrate a substantial increase in trajectory complexity across all measured dimensions following refinement.

\section{Domain Analysis}
\label{app:domain}
\begin{figure}[htbp]
    \centering
    \includegraphics[width=0.5\textwidth]{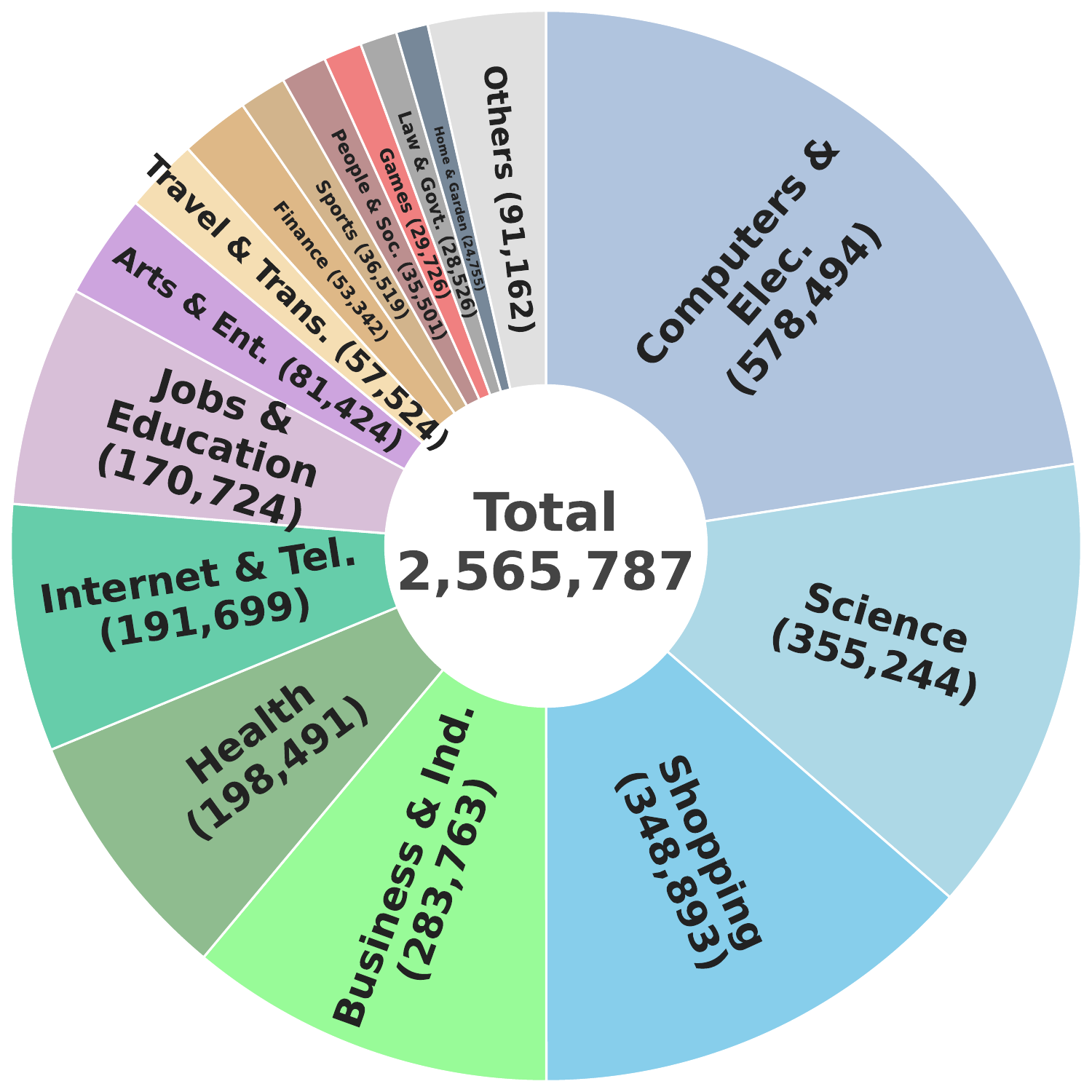}
    \caption{Domain Distribution.}
    \label{fig:tag}
\end{figure}

We analyze the domain distribution in Figure~\ref{fig:tag}. The results demonstrate the diver domains existing in the raw text corpora.

\section{Example of Synthesized Trajectory}
\label{app:full_case}
\begin{figure*}[htbp]
    \centering
    \includegraphics[width=\textwidth]{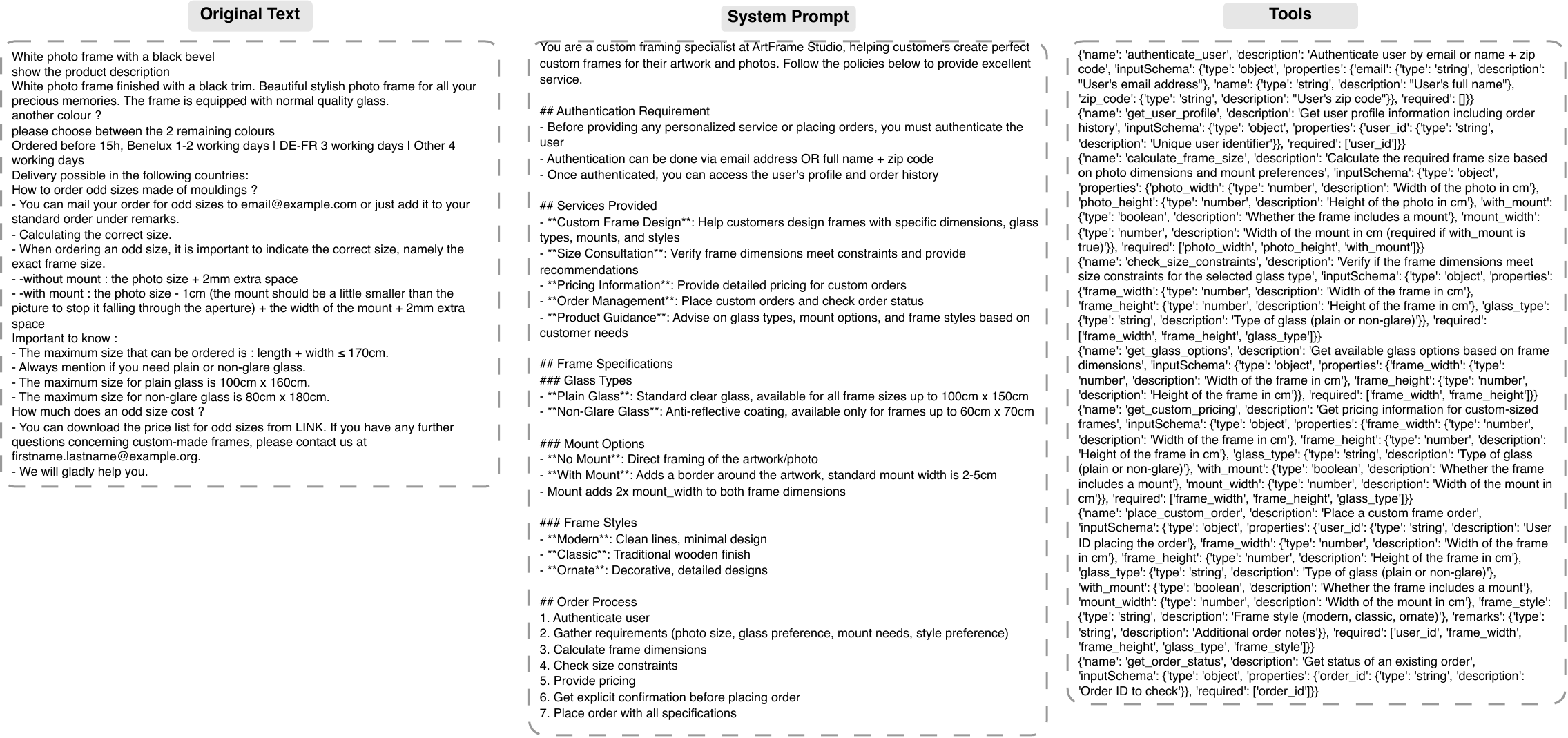}
    \caption{Full example of our generated trajectory (Part 1).}
    \label{fig:full_case_p1}
\end{figure*}

\begin{figure*}[htbp]
    \centering
    \includegraphics[width=\textwidth]{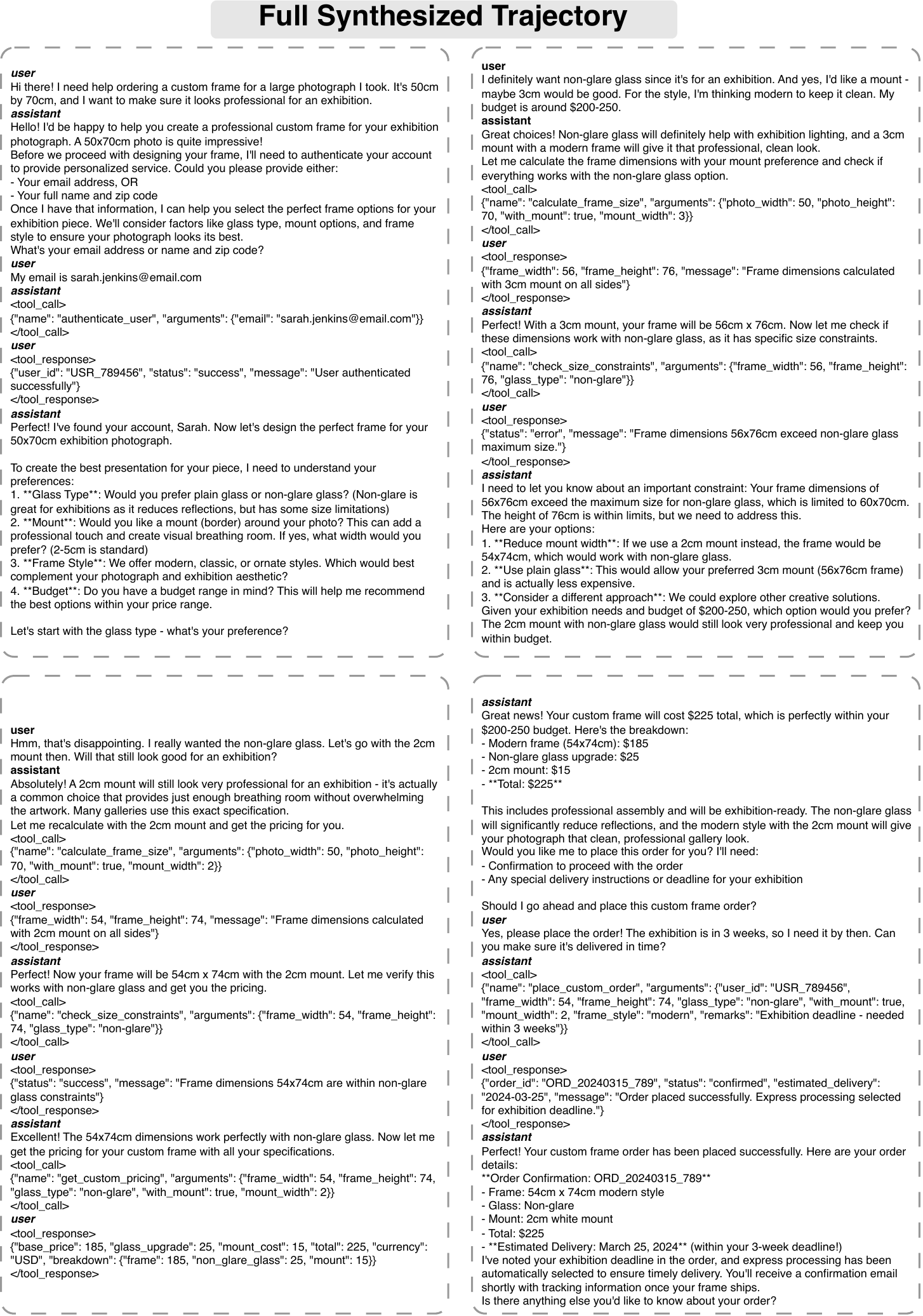}
    \caption{Full example of our generated trajectory (Part 2).}
    \label{fig:full_case_p2}
\end{figure*}

Please refer to Figure~\ref{fig:full_case_p1} and~\ref{fig:full_case_p2}.
The original text provides unstructured information regarding photo frame products, measurement methods, and specific size constraints. The system prompt establishes the persona of a custom framing specialist, enforcing a strict business workflow from user authentication to final order placement. A set of functional tools was designed to support this, including authentication, dimension calculation, constraint checking, and order processing. The final trajectory demonstrates the assistant's exceptional tool-calling and logical reasoning capabilities, particularly its ability to adapt by adjusting parameters and replanning the tool chain after an initial constraint violation, ultimately ensuring a successful transaction.

\end{document}